\documentclass[submission,copyright,creativecommons]{eptcs}
\providecommand{\event}{ICLP 2021} 
\usepackage{breakurl}             
\usepackage{underscore}           

\usepackage{amsmath}
\usepackage{graphicx}
\usepackage{multirow}

\usepackage{graphicx}
\usepackage{latexsym}
\usepackage{amsmath}
\usepackage{blindtext}
\usepackage{algorithmic}
\usepackage[ruled,vlined,linesnumbered,nokwfunc,titlenotnumbered]{algorithm2e}

\newtheorem{defi}{Definition}[section]
\newtheorem{lemma}[defi]{Lemma}
\newtheorem{theorem}[defi]{Theorem}
\newtheorem{cor}[defi]{Corollary}
\newtheorem{ex}[defi]{Example}
\newtheorem{pro}[defi]{Proposition}

\newcounter{myexamplec}
\newenvironment{myexample}{\refstepcounter{myexamplec} \medskip 
\noindent {\bf Example.}~}{\hfill$\Box$\medskip}

\newcommand{\Reduce}{\texttt{Reduce}}

\newcommand{\newc}{\newcommand}

\newc{\comment}[1]{}

\newc{\ap}{\mbox{$\mbox{acyc}_p$}}
\newc{\api}{\mbox{$\mbox{acyc}_{p_i}$}}
\newc{\apij}{\mbox{$\mbox{acyc}_{p_i}^j$}}
\newc{\bpos}{\mbox{$B_{\mbox{pos}}$}}
\newc{\bnbd}{\mbox{$B_{\mbox{neg}}$}}
\newc{\btpos}{\mbox{${B'}_{\mbox{pos}}$}}
\newc{\btnbd}{\mbox{${B'}_{\mbox{neg}}$}}
\newc{\bo}[1]{{\it body}(#1)}
\newc{\card}[1]{\mbox{$|\mbox{#1}|$}}
\newc{\ccd}{\mbox{$\mbox{concl}(\delta)$}}
\newc{\cd}{\mbox{$\mbox{concl}(\delta_1)$}}
\newc{\ccdi}{\mbox{$\mbox{concl}(\delta_i)$}}
\newc{\cb}{\mbox{$\mbox{concl}(\delta_{i-1})$}}
\newc{\cdl}{\mbox{$\mbox{CLOUSES}(\dl)$}}
\newc{\cdn}{\mbox{$\mbox{concl}(\delta_n)$}}
\newc{\ci}{\mbox{Circus}}
\newc{\cp}{\mbox{$C_{P}$}}
\newc{\cnp}{\mbox{$C_{\neg P}$}}
\newc{\cnpc}{\mbox{co-NP-complete}}
\newc{\comp}{\mbox{$\circ$}}
\newc{\concl}{\mbox{$\mbox{concl}$}}
\newc{\ccr}{\mbox{\it clause}}
\newc{\db}{\mbox{$\mbox{\it DB}$}}
\newc{\di}{\mbox{$\delta_i$}}
\newc{\dl}{\mbox{$(D,W)$ \vspace{0.5em}}}
\newc{\dldt}{\mbox{$(\tag{D},W)$}}
\newc{\dlt}{\mbox{$(\tag{D},\tag{W})$}}
\newc{\dn}{\mbox{$\delta_n$}}
\newc{\dol}{\mbox{Dolphine(x)}}
\newc{\dr}{\mbox{$\alpha : \beta / \gamma$}}
\newc{\dt}{\mbox{$\tag{D}$}}
\newc{\du}{\mbox{Dumbo}}
\newc{\ddg}{\mbox{$DDG_{\Pi}$}}  
\newc{\dg}{\mbox{$G_{\Pi}$}} 
\newc{\delt}{\mbox{$\tag{\Delta}$}}   
\newc{\dm}{\mbox{$\triangle M$}}

\newc{\e}{\mbox{$E \hspace{0.2em}$}}
\newc{\eaf}{\mbox{${\mbox{EA}}_{\cal{F}}$}}  
\newc{\edge}{\mbox{$\longrightarrow$}}
\newc{\el}{\mbox{Elephant}}
\newc{\emptyc}{\mbox{$\Lambda$}}
\newc{\en}{\mbox{$E_n$}}
\newc{\entails}{\mbox{$\models$}}
\newc{\es}{\mbox{$E^*$}}
\newc{\etal}{\mbox{ et al.\ }}
\newc{\ext}{\mbox{$\th{E}$}}     
\newc{\eedge}{\mbox{$\stackrel{e}{\longrightarrow}$}}

\newc{\false}{\mbox{\bf false}}
\newc{\fl}{\mbox{Fly}}
\newc{\forest}[1]{\mbox{forest($#1$)}}
\newc{\gp}{\mbox{$G_T$}}
\newc{\grt}{\mbox{$\mbox{ground}_T$}}
\newc{\gs}{\mbox{$G^*$}}  
\newc{\graph}{\mbox{$G(V,E)$}}

\newc{\head}[1]{{\it head}(#1)}
\newc{\h}[1]{{\it head}(#1)}
\newc{\headl}[2]{\mbox{head(#1,#2)}}
\newc{\headd}{\mbox{head($\delta$)}}
\newc{\headdl}[1]{\mbox{head($\delta,#1$)}}
\newc{\headout}[1]{\mbox{head-out(#1)}}
\newc{\headlout}[2]{\mbox{head-out(#1,#2)}}
\newc{\headdout}{\mbox{head-out($\delta$)}}
\newc{\headdlout}[1]{\mbox{head-out($\delta,#1$)}}
\newc{\headdloutp}{\mbox{head-out$^{+}$($\delta,R$)}}  
\newc{\herb}{\mbox{$\cal H$}}
\newc{\ie}{\mbox{i.\ e.\ }}
\newc{\impl}{\mbox{$ \hspace{0.1em} \Rightarrow \hspace{0.1em}$}}
\newc{\inp}{\mbox{$I_{\neg P}$}}
\newc{\ip}{\mbox{$I_{P}$}}
\newc{\ipp}{\mbox{$I_{p}$}}
\newc{\jd}{\mbox{$\mbox{just}(\delta)$}}
\newc{\jdi}{\mbox{$\mbox{just}(\delta_i)$}}
\newc{\jdn}{\mbox{$\mbox{just}(\delta_n)$}}
\newc{\just}{\mbox{$\mbox{just}$}}
\newc{\kb}{\mbox{$\mbox{\it KB}$}}
\newc{\kbe}{\mbox{$\mbox{\it KB}_E$}}
\newc{\kbi}{\mbox{$\mbox{\it KB}_I$}}

\newc{\lang}{\mbox{$\cal L$}}
\newc{\lc}[1]{\mbox{$#1^*$}}
\newc{\ldl}{\mbox{$\lang_{\dl}$}}
\newc{\ldlt}{\mbox{$\tag{\lang}_{\dl}$}}
\newc{\Let}[1]{\mbox{Lett($#1$)}}
\newc{\limp}{\mbox{$\hspace{0.3em}\supset \hspace{0.3em}$}}
\newc{\limm}{\mbox{$\longleftrightarrow$}}
\newc{\lpimp}{\mbox{$\: \longleftarrow \: $}}
\newc{\lpneg}{\mbox{{\it not} \vspace{0.5em}}}
\newc{\lpor}{\mbox{\vspace{0.5em} $\mid$ \vspace{0.5em}}}
\newc{\lit}{\mbox{$\lang$}}
\newc{\liv}{\mbox{Lives-on-land(x)}}
\newc{\lp}{\mbox{$\lang^+$}}
\newc{\lt}{\mbox{${\cal L}_{\Pi}$}}
\newc{\ltt}{\mbox{$\tagg{\lang}$}}
\newc{\m}{\mbox{$\cal M$}}
\newc{\mmp}{\mbox{${\cal M}^3$-Property}}
\newc{\mol}{\mbox{$\overline{M}$}}
\newc{\mtag}{\mbox{$\tag{M}$}}
\newc{\nat}{\mbox{$\cal N$}}
\newc{\nbd}{\mbox{$\lpneg$}}
\newc{\negg}{\mbox{$\sim$}}
\newc{\notsubseteq}{\mbox{$\subseteq \hspace{-0.8em} / \hspace{0.15em} $}}
\newc{\npc}{\mbox{NP-complete}}
\newc{\ncomp}{\mbox{$\comp \hspace{-0.5em} \setminus$}}
\newc{\nmodels}{\mbox{$\models \hspace{-0.8em} \setminus$}}
\newc{\nsmodels}{\mbox{$\smodels \hspace{-0.8em} \setminus$}}
\newc{\nhead}[1]{\mbox{name-head(#1)}}
\newc{\nheadl}[2]{\mbox{name-head(#1,#2)}}
\newc{\nheadd}{\mbox{name-head($\delta$)}}
\newc{\nheaddl}[1]{\mbox{name-head($\delta,#1$)}}
\newc{\nheadout}[1]{\mbox{name-head-out(#1)}}
\newc{\nheadlout}[2]{\mbox{name-head-out(#1,#2)}}
\newc{\nheaddout}{\mbox{name-head-out($\delta$)}}
\newc{\nheaddlout}[1]{\mbox{name-head-out($\delta,#1$)}}  

\newc{\order}{\mbox{$\prec$}}
\newc{\orr}{\mbox{$\:\mid\:$}}   
\newc{\p}{\mbox{${\cal P}_{\dl}$}}
\newc{\pdi}{\mbox{$\mbox{pre}(\delta_i)$}}
\newc{\pdn}{\mbox{$\mbox{pre}(\delta_n)$}}
\newc{\pkb}{\mbox{$\mbox{\it PKB}$}}
\newc{\pim}[1]{\mbox{$PI(#1)$}}
\newc{\piw}{\mbox{$\mbox{$W^+$}$}}
\newc{\pnp}{\mbox{$\mbox{$P^{NP[O(\log n)]}$}$}}
\newc{\ppd}{\mbox{$\mbox{pre}(\delta)$}}
\newc{\ppt}{\mbox{$\Pi^P_2$}}
\newc{\pr}{\mbox{{\em Proof: }}}
\newc{\pre}{\mbox{$\mbox{pre}$}}
\newc{\prov}{\mbox{$\vdash$}}
\newc{\piord}{\mbox{$\Pi_{ord}$}}
\newc{\piweek}{\mbox{$\Pi_{week}$}}

\newc{\quoted}[1]{\mbox{``#1"}}
\newc{\rel}{\mbox{$\cal R$}}
\newc{\res}[2]{\mbox{$res(#1,#2)$}}
\newc{\refi}[1]{\mbox{(\ref{#1})}}
\newc{\rg}{\mbox{$RG_{\Pi}$}}
\newc{\rr}{\mbox{\it rel}}
\newc{\scc}[1]{\mbox{scc($#1$)}}
\newc{\smodels}{| \hspace{-0.45em} \approx}
\newc{\sptc}{\mbox{$\Sigma^p_2$-complete}}
\newc{\spth}{\mbox{$\Sigma^p_2$-hard}}
\newc{\srg}{\mbox{$SRG_{\Pi}$}}
\newc{\st}{\mbox{\tag{S}}}     
\newc{\sddg}{\mbox{$SDDG_{\Pi}$}}
\newc{\sg}{\mbox{$\cal G$}}
\newc{\sv}{\mbox{$\cal V$}}
\newc{\se}{\mbox{$\cal E$}}   
\newc{\sstr}{\mbox{$\sg(\sv, \se)$}}

\newc{\tagg}[1]{\mbox{$#1^{\prime \prime}$}}
\newc{\td}{\mbox{$\tag{\delta}$}}
\newc{\te}{\mbox{$\tag{E}$}}
\newc{\tlt}{\mbox{$\tag{\lt}$}}
\newc{\tl}{\mbox{$\tag{L}$}}
\newc{\tm}{\mbox{$\tag{M}$}}
\newc{\tp}{\mbox{$T_\Pi$}}
\newc{\ttp}{\mbox{$\tag{T}_\Pi$}}
\newc{\tree}[1]{\mbox{tree($#1$)}}
\newc{\true}{\mbox{\bf true}}
\newc{\union}{\mbox{$\bigcup$}}
\newc{\unions}{\mbox{$\bigcup_*$}}
\newc{\vrg}{\mbox{$VRG_{\Pi}$}}
\newc{\vs}{\mbox{$V^*$}}        
\newc{\vsrg}{\mbox{$VSRG_{\Pi}$}}

\newc{\wa}{\mbox{Warm-blooded(x)}}
\newc{\w}{\mbox{$w^*$}}
\newc{\wt}{\mbox{$\tag{W}}}
\newc{\wits}[2]{\mbox{$Wit_{#1}(#2)$}}
\newc{\witl}[3]{\mbox{$Wit_{#1,#2}(#3)$}}

\title{
How to Split a Logic Program}

\author{Rachel Ben-Eliyahu-Zohary
\institute{Department of Software Engineering \\ JCE- Azrieli College of Engineering \\ Jerusalem, Israel}
\email{rbz@jce.ac.il}
}

\def\titlerunning{How to Split a Logic Program}
\def\authorrunning{R. Ben-Eliyahu-Zohary}
\begin{document}
\maketitle

\begin{abstract}
Answer Set Programming (ASP) is a successful method for solving a range of real-world applications. Despite the availability of fast ASP solvers, computing answer sets demands a very large computational power, since the problem tackled is in the second level of the polynomial hierarchy. A speed-up in answer set computation may be attained, if the program can be split into two disjoint parts, bottom and top. Thus, the bottom part is evaluated independently of the top part, and the results of the bottom part evaluation are used to simplify the top part. Lifschitz and Turner have introduced the concept of a splitting set, i.e., a set of atoms that defines the splitting. 

In this paper, We show that the problem of computing a splitting set with some desirable properties can be reduced to a classic Search Problem and solved in polynomial time. This allows us to conduct experiments on the size of the splitting set in various programs and lead to an interesting discoery of a source of complication in stable model computation.
We also show that for Head-Cycle-Free programs, the definition of splitting sets can be adjusted to allow splitting of a broader class of programs. 
\end{abstract}
%

\section{Introduction}\label{sect:intro}
Answer Set Programming (ASP) is a successful method for solving a range of real-world applications. Despite the availability of fast ASP solvers, the task of computing answer sets demands extensive computational power, because the problem tackled is in the second level of the polynomial hierarchy. A speed-up in answer set computation may be gained, if the program can be divided into several modules in which each module is computed separately \cite{LiTu94,JOTW09,FLLP09}. Lifschitz and Turner propose to split a logic program into two disjoint parts, bottom and top, such that the bottom part is evaluated independently from the top part, and the results of the bottom part evaluation are used to simplify the top part. They have introduced the concept of a splitting set, i.e., a set of atoms that defines the splitting \cite{LiTu94}. 
In addition to inspiring incremental ASP solvers \cite{GKKOST08}, splitting sets are shown to be useful also in investigating answer set semantics \cite{DEFK09,OiJa08,FLLP09}. 

In this paper
we raise and answer two questions
regarding splitting sets.
The first question is, how do we compute a splitting set? We show that if we are looking for a splitting set having a desirable property that can be tested efficiently, we can find it in polynomial time. Examples of desirable splitting sets can be minimum-size splitting sets, splitting sets that include certain atoms, or splitting sets that define a bottom part with minimum number of rules or bottom that are easy to compute, for example, a bottom which is an HCF program \cite{BeDe94}. Once we have an effcient algorithm for computing splitting sets, we can use it to investigate the size of a minimal noempty splitting set and discover some interesting results that explain the source of complexity in computing stable models.

Second, we ask if it is possible to relax the definition of splitting sets such that we can now split programs that could not be split using the original definition. We answer affirmatively to the second question as well, and we present a more general and relaxed definition of a splitting set.


\section{Preliminaries}\label{sect:preliminaries}

\subsection{Disjunctive Logic Programs and Stable Models}
A propositional {\em Disjunctive Logic Program} (DLP) is
a collection of rules of the form
\begin{eqnarray*}
A_1 | \ldots | A_k \lpimp A_{k+1}, \ldots, A_m, \nbd A_{m+1},
\ldots, \nbd A_n, \hspace{0.5 cm}
n\ge m\ge k\ge 0,
\end{eqnarray*}
where the symbol $\quoted{\nbd}$ denotes negation by default, and
each $A_i$ is an atom (or variable). 
For $k+1 \leq i \leq m$, we will say that $A_i$ appears {\em positive} in the body of the rule, while for $m+1 \leq i \leq n$, we shall say that $A_i$ appears {\em negative} in the body of the rule.
If $k=0$, then the rule
is called {\em an integrity rule}. If $k > 1$, then the rule
is called {\em a disjunctive rule}.
The expression to the left of $\lpimp$ is called the {\em head}
of the rule, while the expression to the right of $\lpimp$ is
called the
{\em body} of the rule. Given a rule $r$, $\h{r}$ denotes the set of atoms in the head of $r$, and $\bo{r}$ denotes the set of atoms in the body
of $r$.
We shall shall sometimes denote a rule by $H \lpimp \bpos, \bnbd$, 
where $\bpos$ is the set of positive atoms in the body of the rule ($A_{k+1},...,A_m$), $\bnbd$ is the set of negated atoms in the body of the rule ($A_{m+1},...,A_n$), and $H$ the set of atoms 
in its head. Given a program ${\cal{P}}$, $\Let{\cal{P}}$ is the set of all atoms that appear in ${\cal{P}}$.
From now, when we refer to a program, it is a DLP.

Stable Models \cite{GeLi91} of a program $\cal{P}$ are defined as
Follows: Let $\Let{\cal{P}}$ denote the set of all atoms occurring in $\cal{P}$. Let a {\em
context} be any subset of $\Let{\cal{P}}$. Let $\cal{P}$ be a
{\em negation-by-default-free} program. Call a context $S$ {\em
closed under $\cal{P}$} iff for each rule $A_1 | \ldots | A_k \leftarrow
A_{k+1}, \ldots, A_m$ in $\cal{P}$, if $A_{k+1}, \ldots, A_m \in S$,
then for some $i=1,\ldots,k$, $A_i \in S$. A {Stable Model} of
$\cal{P}$ is any minimal context $S$, such that $S$ is closed under
$\cal{P}$ . A stable model of a general
DLP is defined as follows: Let the {\em reduct of $\cal{P}$ w.r.t. $\cal{P}$ and the
context $S$} 
be the DLP obtained from
$\cal{P}$ by deleting ($i$) each rule that has $\nbd A$ in its body for
some $A \in S$, and ($ii$) all subformulae of the form $\nbd A$ of
the bodies of the remaining rules. Any context $S$ which is a stable model
of the reduct of $\cal{P}$ w.r.t. $\cal{P}$ and the
context $S$ is a {\em stable model} of $\cal{P}$. 

HCF- 
\textit{Head Cycle Free} programs \cite{BeDe94}
are DLPs such that 
in the associated dependency graph there is
no cycle including
two atoms 
occurring in the head of the same rule.


  \begin{defi} A set of atoms $S$ {\em  satisfies} the body of a rule $r$ if all the
   atoms that appear positive in the body of $r$ are in $S$ and all the atoms that appear negative in $r$ are not in $S$. A set of atoms $S$ {\em  satisfies}  a rule if it does not {\em  satisfy} the body of the rule $r$ or if one of the atoms in $\h{r}$ is in $S$.
  \end{defi} 
According to \cite{BeDe94}, a proof of a atom is a sequence of rules that can
   be used to derive the atom from the program. 
 \begin{defi}[ \cite{BeDe94}]
\label{def:proof}
  An atom $L$ has a {\em proof } w.r.t. a set of atoms $S$ and a logic program ${\cal{P}}$  if and only if there is a sequence of rules $r_1,...,r_n$ from ${\cal{P}}$ such that: 

\begin{enumerate}
 \item for all $1 \leq i \leq n$ there is one and only one atom in the head of $r_i$ that belongs to $S$.
  \item  $L$ is the head of $r_n$. \item  for all $1 \leq i \leq n$, the body of $r_i$ is satisfied by $S$. \item  $r_1$ has no atoms that appear positive in its body, and for each $1 < i \leq n$, each atom that appears positive in the body of $r_i$ is in the head of some $r_j$ for some $1 \leq j < i$.
      \end{enumerate}

 \end{defi}
Note that given a proof $r_1,...,r_n$ of some atom $L$  w.r.t. a set of atoms $S$ and a logic program ${\cal{P}}$, for every $1 \leq i \leq n$ $r_1,...,r_i$ is also a proof of some atom $L'$  w.r.t.  $S$ and  ${\cal{P}}$.

\begin{theorem}[\cite{BeDe94}] Let ${\cal{P}}$ be an HCF DLP. Then a set of atoms $S$  is an answer set of ${\cal{P}}$ if and only if $S$ satisfies 
 each rule in ${\cal{P}}$ and  each $a \in S$ has a proof with respect to ${\cal{P}}$ and $S$. \end{theorem}

\subsection{Programs and graphs}
With every program $\cal{P}$ we associate a directed graph,
called the {\em dependency
graph} of $\cal{P}$,
in which (a) each atom in $\Let{\cal{P}}$ is a node,
and (b) 
there is an arc directed from a node $A$ to a node $B$ if there is a rule $r$ in $\cal{P}$ such that $A \in \bo{r}$ and $B \in \h{r}$.

\begin{figure}
\centering
\includegraphics[width=\linewidth]{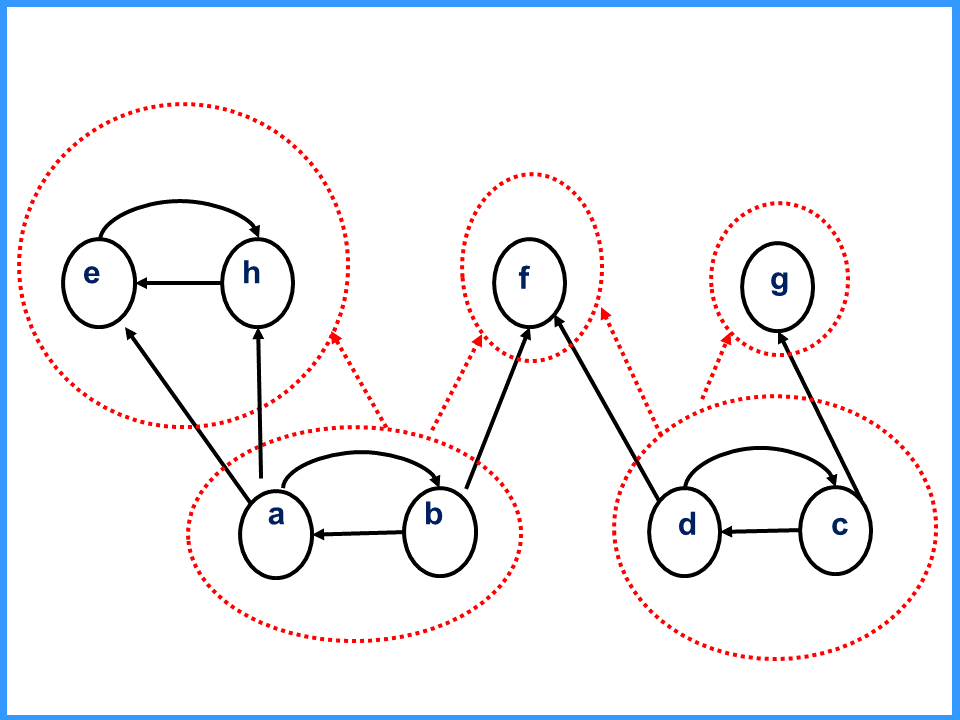}

\caption{The [super]dependency graph of the program $\cal{P}$.}
\label{dgraph}
\end{figure}

A \textit{super-dependency graph} $SG$ is an acyclic graph 
built from a dependency graph $G$ as follows:
For each strongly connected component (SCC) $c$ in $G$ there is a node in $SG$,
and for each arc in $G$ from a node in a strongly connected component $c_1$ to a 
node in a strongly connected component $c_2$ (where $c_1 \neq c_2$) there is an arc in $SG$ from the 
node associated with $c_1$ to the node associated with $c_2$.
A program $\cal{P}$ is Head-Cycle-Free (HCF), if there are no two atoms in the head of some rule in $\cal{P}$ that belong to the same component in the super-dependency graph of $\cal{P}$ \cite{BeDe94}.
Let $G$ be a directed graph and $SG$ be a super dependency graph of $G$. A {\em source} in $G$ (or $SG$) is a node with no incoming edges.
By abuse of terminology, we shall sometimes use the term ``source" or ``SCC" as the set of nodes in a certain source or a certain SCC in $SG$, respectively, and when there is no possibility of confusion we shall use the term rule for the set of all atoms that appears in the rule. Given a node $v$ in $G$, $\scc{v}$ denotes the set of all nodes in the SCC in $SG$ to which $v$ belongs, and $\tree{v}$ denotes the set of all nodes that belongs to any SCC $S$ such that there is a path in $SG$ from $S$ to $\scc{v}$. Similarly, when $S$ is a set of nodes, $\tree{S}$ is the union of all $\tree{v}$ for every $v \in S$. 
 Given a node $v$ in $G$, $\scc{v}$ will be sometimes called the {\em root } of $\tree{v}$.
For example, given the super dependency graph in Figure \ref{dgraph}, $\scc{e}=\{e,h\}$, $\tree{e}=\{a,b,e,h\}$, $\tree{\{f,g\}}=\{a,b,c,d,f,g\}$ and $\tree{r}$, where $r=$ $c| f \lpimp \nbd d$ is actually $\tree{\{c,d,f\}}$ which is $\{a,b,c,d,f\}$.

{\em A source in a program } will serve as a shorthand for “a source in the super dependency graph of the program.”
Given a source $S$ of a program $\cal{P}$, $\cal{P}_S$ denotes the set of rules in $\cal{P}$ that uses only atoms from $S$.

\begin{ex}[Running Example]
\label{re}
Suppose we are given the following program $\cal{P}$:
\{
1.   $a$   $\lpimp$   $\nbd b$  ,
2.   $e | b $   $\lpimp$   $\nbd a$  ,
3.   $f$   $\lpimp$   $\nbd b$ ,
4.   $g | d$ $\lpimp$   $c$  ,
5.   $c|f $  $\lpimp$   $\nbd d$  ,
6.   $h$  $\lpimp$   $e$  ,
7.   $e$  $\lpimp$   $a, \nbd h$  ,
8.   $h$   $\lpimp$   $a$ 
\}
In Figure 1 the dependency graph of $\cal{P}$ is illustrated in {\bf solid} lines. The SG is marked with {\bf dotted} lines. Note that $\{a,b\}$ is a source in the SG of $\cal{P}$, but it is not a splitting set.
\end{ex}
\subsection{Splitting Sets}

The definitions of {\em Splitting Set} and the {\em Splitting Set Theorem} are adopted from a paper by Lifschitz and Turner \cite{LiTu94}. We restate them here using the notation and the limited form of programs discussed in our work.
\begin{defi}[Splitting Set] A {\em Splitting Set} for a program $\cal{P}$ is a set of of atoms $U$ such that for each rule $r$ in $\cal{P}$, if one of the atoms in the head of $r$ is in $U$, then all the atoms in $r$ are in $U$. We denote by $b_U(\cal{P})$ the set of all rules in $\cal{P}$ having only atoms from $U$.
\end{defi}

The empty set is a splitting set for any program.
For an example of a nontrivial splitting set, the set $\{a,b,e,h\}$ is a splitting set for the program $\cal{P}$ introduced in Example \ref{re}. The set $b_{\{a,b,e,h\}}(\cal{P})$ is $\{r_1,r_2,r_6,r_7,r_8 \}$.

For the Splitting set theorem, we need the a procedure called $\mbox{Reduce}$, which resembles many reasoning methods in knowledge representation, as, for example, unit propagation in DPLL and other constraint satisfaction algorithms
\cite{DLL62,Dec03}. 
\mbox{Reduce}($\cal{P}$,$X$,$Y$) returns the program obtained from a given program $\cal{P}$ in which all atoms in $X$ are set to true,
and all atoms in $Y$ are set to false. \mbox{Reduce}($\cal{P}$,$X$,$Y$) is shown in Figure \ref{proc:reduce}. For example, 
\mbox{Reduce}($\cal{P}$,$\{a,e,h\}$,$\{b\}$), where $\cal{P}$ is the program from Example \ref{re}, is the following program (the numbers of the rules are the same as the  corresponding rules of the program in Example \ref{re}):
\{
3.  $f$ $\lpimp$,
4.  $g | d$ $\lpimp$  $c$,
5.  $c|f $ $\lpimp$  $\nbd d$
\}

\begin{procedure}[t]
\small
\KwIn{A program $\cal{P}$ and two sets of atoms: $X$ and $Y$}
\KwOut{An update of $\cal{P}$ assuming all the atoms in $X$ are true and all atoms in $Y$ are false}
\BlankLine
\ForEach{{\rm atom} $a \in X$}
{
\ForEach{ rule $r$ in $\cal{P}$}{

If $a$ appears negative in the body of $r$ delete $r$ \;
If $a$ is in the head of $r$ delete $r$\;
Delete each positive appearance of $a$ in the body of $r$\;
}
}
\ForEach{{\rm atom} $a \in Y$}
{
\ForEach{ rule $r$ in $\cal{P}$}{

If $a$ appears positive in the body of $r$, delete $r$ \;
If $a$ is in the head of $r$, delete $a$ from the head of $r$\;
Delete each negative appearance of $a$ in the body of $r$\;
}
}

return $\cal{P}$\;
\caption{\mbox{Reduce}($\cal{P}$,$X$,$Y$)}
\label{proc:reduce}
\end{procedure}

\begin{theorem}[Splitting Set Theorem] (adopted from \cite{LiTu94})
Let $\cal{P}$ be a program, and let $U$ be a splitting set for $\cal{P}$. A set of atoms $S$ is a stable model for $\cal{P}$ if and only if $S=X \cup Y$, where $X$ is a stable model of 
$b_U(\cal{P})$, and $Y$ is a stable of $\mbox{Reduce}({\cal{P}},X,U-X)$.
\end{theorem}

As seen in Example \ref{re}, a source is not necessarily a splitting set. A slightly different definition of a dependency graph is possible. The nodes are the same as in our definition, but in addition to the edges that we already have, we add a directed arc from a variable $A$ to a variable $B$ whenever $A$ and $B$ are in the head of the same rule. It is clear that a source in this variation of dependency graph must be a splitting set. The problem is that the size of a dependency graph built using this new definition may be exponential in the size of the head of the rules, while we are looking for a polynomial-time algorithm for computing a nontrivial splitting set.

\subsection{Search Problems}
\begin{figure}
\includegraphics[width=\linewidth]{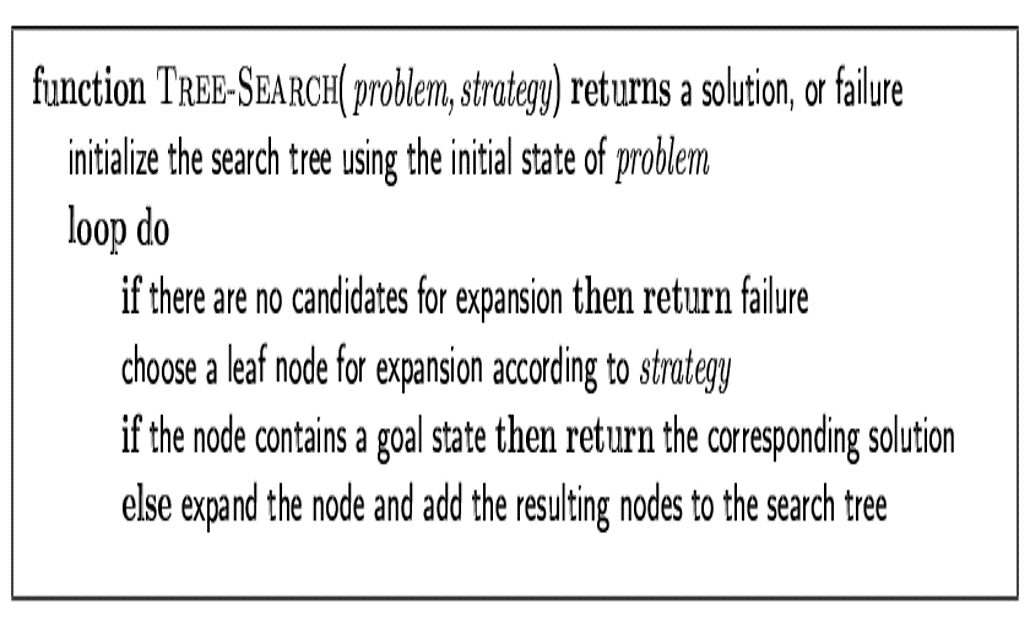}
\caption{Tree Search Algorithm}
\label{tree}
\end{figure}
The area of {\em search} is one of the most studied and most known areas in AI (see, for example, \cite{Pea84}). In this paper we show how the problem of computing a nontrivial minimum-size splitting set can be expressed as a search problem. We first recall basic definitions in the area of {\em search}.
A {\em search problem} is defined by five elements:
set of states, initial state, actions or successor function, goal test, and path cost.
\comment{
A search problem over a state space problem is defined by five items:
\begin{enumerate}
\item set of states
\item initial state
\item actions or successor function
\item goal test
\item path cost (additive)
\end{enumerate}
}
A {\em solution} is a sequence of actions leading from the initial state to a goal state. Figure \ref{tree} provides a basic search algorithm \cite{RuNo10}.

There are many different strategies to employ when we choose the next leaf node to expand. In this paper we use {\em uniform cost}, according to which we expand the leaf node with the lowest path cost.

\section{Between Splitting Sets and Dependency Graphs}\label{spdg}

In this section we show that a splitting set is actually a tree in the SG of the program $\cal{P}$.
The first lemma states that if an atom $Q$ is in some splitting set, all the atoms in $\scc{Q}$ must be in that splitting set as well.
\begin{lemma}\label{hs}
Let $\cal{P}$ be a program, let $SP$ be a Splitting Set in $\cal{P}$, let $Q \in SP$, and let $S=\scc{Q}$. It must be the case that $S \subseteq SP$.
\end{lemma}
{\em{Proof:}}
Let $R \in S$. We will show that $R \in SP$.
Since $Q \in S$, and $S$ is a strongly connected component, it must be that for each $Q' \in S$ there is a path in $SG$ -the super dependency graph of ${\cal{P}}$ - from $Q'$ to $Q$, such that all the atoms along the path belong to $S$. The proof goes by induction on $i$, the number of edges in the shortest path from $Q'$ to $Q$.
\begin{description}
\item[Case $i=0$.] Then $Q=Q'$, and so obviously $Q' \in SP$.
\item[Induction Step.] Suppose that for all atoms $Q' \in S$, such that the shortest path from $Q'$ to $Q$ is of size $i$, $Q'$ belongs to $SP$. Let $R$ be an atom in $S$, such that the shortest path from $R$ to $Q$ is of size $i+1$. So, there must be an atom $R'$ such that there is an edge in $SG$ from $R$ to $R'$, and the shortest path from $R'$ to $Q$ is of size $i$. By the induction hypothesis, $R' \in SP$. Since there is an edge from $R$ to $R'$ in $SG$, it must be that there is a rule $r$ in $\cal{P}$, such that $R \in\bo{ r}$ and $R' \in \h{r}$. Since $R' \in SP$ and $SP$ is a Splitting Set, it must be the case that $R \in SP$.
\end{description}

\begin{lemma}
Let $\cal{P}$ be a program, let $SP$ be a Splitting Set in $\cal{P}$, let $r$ be a rule in $\cal{P}$, and $S$ an SCC in $SG$ -- the super dependency graph of ${\cal{P}}$. If $\h{r} \cap SP \neq \emptyset$, then 
$\tree{r} \subseteq SP$.
\end{lemma}
{\em{Proof:}}
We will show that for every $Q \in r$, $\tree{Q} \subseteq SP$. Let $Q \in r$. The set $\tree{Q}$ is a union of SCCs. We shall show that for every SCC $S$ such that $S \subseteq \tree{Q}$, $S \subseteq SP$.
Let $S'$ be the root of $\tree{Q}$. The proof is by induction on the distance $i$ from $S$ to $S'$.
\begin{description}
\item[Case $i=0$.] Then $S=S'$, and  $S$ is the root of $\tree{Q}$.  Since $\h{r} \cap SP \neq \emptyset$, $Q \in r$ and $SP$ is a splitting set, $Q \in SP$. So by Lemma \ref{hs} $S \subseteq SP$.
\item[Induction Step.] Suppose that for all SCCs $S \in \tree{Q}$ such that the distance from $S$ to $S'$ is of size $i$ $S \subseteq SP$. Let $R$ be an SCC in $\tree{r}$, such that the distance from $R$ to $S'$ is of size $i+1$. So, there must be an SCC $R'$, such that there is an edge in $\tree{r}$ from $R$ to $R'$, and the distance from $R'$ to $S'$ is of size $i$. By the induction hypothesis, $R' \subseteq SP$. Since there is an edge from $R$ to $R'$ in $\tree{Q}$, it must be the case that there is a rule $r$ in $\cal{P}$, such that an atom from $R$, say $P$, is in $\bo{ r}$, and an atom from $R'$, say $P'$, is in $\h{r}$. By induction hypothesis, $P' \in SP$, and since $SP$ is a Splitting Set, it must be that $P \in SP$. By Lemma \ref{hs}, $R \subseteq SP$.
\end{description}

\begin{cor} \label{stree} Every Splitting set is a collection of trees.
\end{cor}

Note that the converse of Corollary \ref{stree} does not hold. In our running example, for instance, $\tree{g}=\{c,d,g\}$, but $\{c,d,g\}$ is not a splitting set.

\section{Computing a minimum-size Splitting Set as a search problem}\label{sect:algo}

We shall now confront the problem of computing a splitting set with a desirable property. We shall focus on computing a nontrivial minimum-size splitting set. Given a program $\cal{P}$, this is how we view the task of computing a nontrivial minimum-size splitting set as a search problem. We assume that there is an order over the rules in the program.

\begin{description}
\item[State Space.] The state space is a collection of forests which are subgraphs of the super dependency graph of $\cal{P}$.
\item[Initial State.] The empty set.
\item[Actions.] \begin{enumerate} \item The initial state can unite with one of the sources in the super dependency graph of $\cal{P}$.
\item A state $S$, other than the initial state, has only one possible action, which is: \begin{enumerate} \item Find the lowest rule $r$ (recall that the rules are ordered) such that $\h{r} \cap S \neq \emptyset$ and $\Let{r} \not \subseteq S$; \item Unite $S$ with $\tree{r}$. \end{enumerate}
\end{enumerate}
\item[Transition Model] The result of applying an action on a state $S$ is a state $S'$ that is a superset of $S$ as the actions describe.
\item[Goal Test] A state $S$ is a goal state, if there is no rule $r \in \cal{P}$ such that $\h{r} \cap S \neq \emptyset$ and $\Let{r} \not \subseteq S$. (In other words, a goal state is a state that represents a splitting set.); 
\item[Path Cost] The cost of moving from a state $S$ to a state $S'$ is $|S'| - |S|$, that is the number of atoms added to $S$ when it was transformed to $S'$. So, the path cost is actually the number of atoms in the final state of the path.
\end{description}

Once the problem is formulated as a search problem, we can use any of the search algorithms developed in the AI community to solve it. We do claim here, however, that the computation of a nontrivial minimum-size splitting set can be done in time that is polynomial in the size of the program. This search problem can be solved, for example, by a search algorithm called Uniform Cost. 
Algorithm Uniform Cost \cite{RuNo10} is a variation of Dijkstra's single-source
shortest path algorithm \cite{Dij59,Fel11}. Algorithm Uniform Cost is optimal, that is,  it returns a shortest path to a goal state.
Since the search problem is formulated so that the length of the path to a goal state is the size of the splitting set that the goal state represents, Uniform Cost will find a minimum-size splitting set.

The time complexity of this algorithm is $O(b^m)$, where $b$ is the branching factor of the search tree generated, and $m$ is the depth of the optimal solution. It is easy to see that $m$ cannot be larger than the number of rules in the program, because once we use a rule for computing the next state, this rule cannot be used any longer in any sequel state. As for $b$, the branching factor, except for the initial state, each state can have at most one child; to generate a child we apply the lowest rule that demonstrates that the current state is not a splitting set. In a given a specific state, the time that required to calculate its child is polynomial in the size of the program. Therefore, this search problem can be solved in polynomial time. This claim is summarized in the following proposition.

\begin{pro}
A minimum-size nontrivial splitting set can be computed in time polynomial in the size of the program.
\end{pro}

The following example demonstrates how the search algorithm works, assuming that we are looking for the smallest non-empty splitting set, and we are using uniform cost search.

\begin{figure}[h]
\centering
\includegraphics[width=\linewidth]{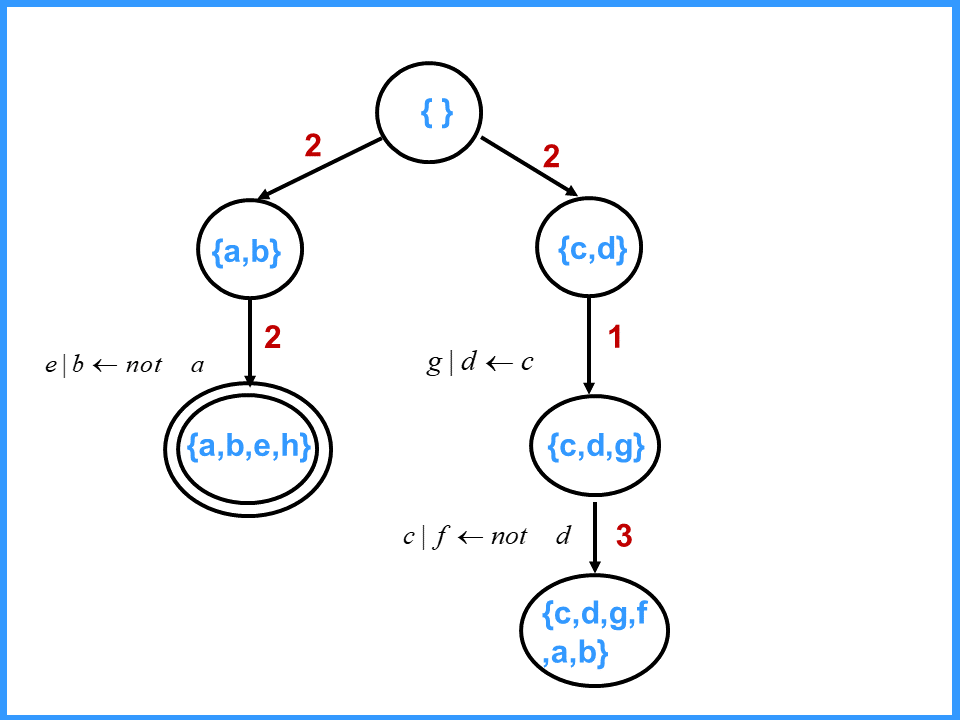}
\caption{The search tree for $\cal{P}$.}
\label{tree1}
\end{figure}

\begin{myexample}
\label{moduex}
Suppose we are given the program $\cal{P}$ of Example \ref{re}, and we want to apply the search procedure to compute a nontrivial minimum-size splitting set. The search tree is shown in Figure \ref{tree1}.
Our initial state is the empty set. By the definition of the search problem, the successors of the empty set are the sources of the super dependency graph of the program, which in this case are $\{a,b\}$ and $\{c,d\}$, both of which with action cost 2. Since both current leaves have the same path cost, we shall choose randomly one of them, say $\{c,d\}$, and check whether it is a goal state, or in other words, a splitting set. It turns out $\{c,d\}$ is not a splitting set, and the lowest rule that proves it is rule No. 4 that requires a splitting set that includes $d$ to have also $c$ and $g$. So, we make the leaf $\{c,d,g\}$ the son of $\{c,d\}$ with action cost 1 (only one atom, $g$, was added to $\{c,d\}$). Now we have two leaves in the search tree. The leaf $\{a,b\}$ with path cost 2, that was there before, and the leaf $\{c,d,g\}$, that was just added, with path cost 3. So we go and check whether $\{a,b\}$ is a splitting set and find out that Rule no. 2 is the lowest rule that proves it is not. W,e add the tree of Rule no. 2 and get the child $\{a,b,e,h\}$ with a path cost 4. So, we go now and check whether $\{c,d,g\}$ is a splitting set and find that Rule no. 5 is the lowest rule that proves that it is not. We add the tree of Rule no. 5 and get the child $\{c,d,g,f,a,b\}$ with a path cost 6. Back to the leaf $\{a,b,e,h\}$, the leaf with the shortest path, we find that it is also a splitting set, and we stop the search.
\end{myexample}

\section{Experiments}
\begin{figure}
\centering
\includegraphics[width=\linewidth]{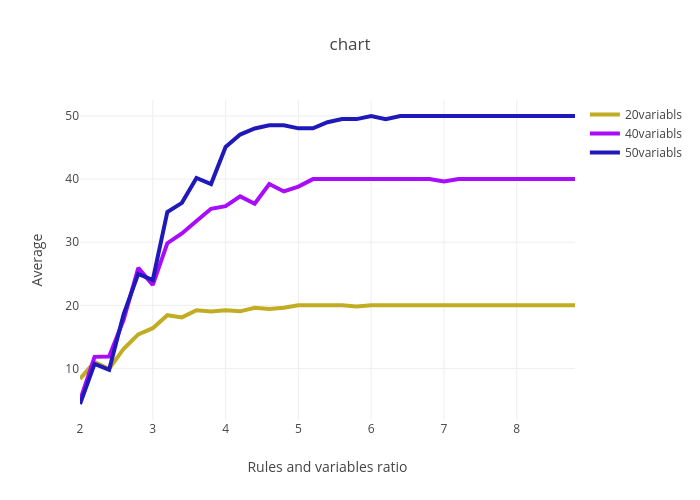}
\caption{Average size of nonempty splitting sets.}
\label{sa}
\end{figure}

\begin{figure}
\centering
\includegraphics[width=\linewidth]{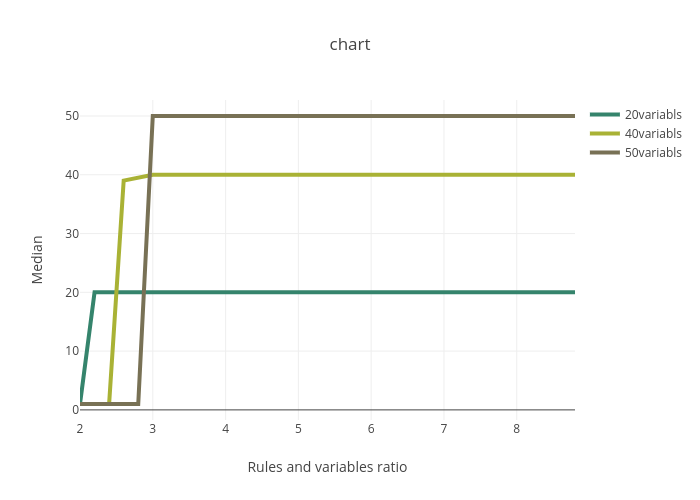}
\caption{Median size of nonempty splitting sets.}
\label{sm}
\end{figure}

We have implemented our algorithm and tested it on randomly generated programs, having no negation as failure. Each rule in the program has exactly 3 variables where any subset of them can be in the head.  A stable model is actually a minimal model for this type of program.
For each program we have computed  a nontrivial minimum-size splitting set.
The average nontrivial minimum size of a splitting set, and the median of all nontrivial minimum size splitting sets, as a function of the rules to variable number ratio, are shown in Graph \ref{sa} and
Graph \ref{sm}, respectively. The average and median were taken over 100 programs generated randomly, starting with a ratio of 2 and generating 100 random programs for each interval of 0.25. It is clear from the graphs that in the transition value of 4.25 (See \cite{SeMiLe96}) the size of the splitting set is maximal, and it is equal to the number of variables in the program.
This is a new way of explaining that, programs in the phase transition value of rules to variable are hard to solve

\section{Relaxing the splitting set condition}
As the experiments indicate, in the hard random problems the only nonempty splitting set is the set of all atoms in the program. In such cases splitting is not useful at all. In this section we introduce the concept of {\em generalized splitting set} (g-splitting set), which is a relaxation of the concept of a splitting set.
Every splitting set is a g-splitting set, but there are g-splitting sets that are not splitting sets. 

\begin{defi}[Generalized Splitting Set.] A {\em Generalized Splitting Set (g-splitting set)} for a program $\cal{P}$ is a set of of atoms $U$ such that for each rule $r$ in $\cal{P}$, if one of the atoms in the head of $r$ is in $U$, then all the atoms in the body of $r$ are in $U$. 
\end{defi}
Thus, g-splitting sets that are not splitting sets may be found only when there are disjunctive rules in the program.
\begin{ex}
\label{eg}
Suppose we are given the following program $\cal{P}$:
\{
1.  $a$  $\lpimp$  $\nbd b$ ,
2.  $b $  $\lpimp$  $\nbd a$ ,
3.  $b | c$ $\lpimp$  $a$ ,
4.  $a | d$ $\lpimp$  $b$ 
\}
The program has only the two trivial splitting sets --- the empty set and $\{a,b,c,d\}$. However, the set $\{a,b\}$ is a g-splitting set of $\cal{P}$.
\end{ex}

We next demonstrate the usefulness of g-splitting sets. We show that it is possible to compute a stable model of an HCF program $\cal{P}$ by computing a stable model of $\cal{P}_S$ for a g-splitting set $S$ of $\cal{P}$, and then propagating the values assigned to atoms in S to the rest of the program.

\begin{theorem}[program decomposition.]\label{gd}
Let $\cal{P}$ be a HCF  program. For any g-splitting-set $S$ in $\cal{P}$, 
let $X$ be a stable model of ${\cal{P}}_S$.
Moreover, let $\cal{P}'=\mbox{Reduce}({\cal{P}},$X$,$S-X$)$, where $\mbox{Reduce}({\cal{P}},X,S-X)$ is the result of propagating the assignments of the model $X$ in the program $\cal{P}$. 
Then,
for any 
stable model $M'$ of $\cal{P}'$, $M' \cup X$ 
is a stable model of $\cal{P}$.
\label{thrm:modumin}
\end{theorem}

{\em{Proof:}}
We denote $M' \cup X$ by $M'X$. 
The proof has two steps. We prove that (1)- 
  $M'X$ is a model of $\cal{P}$ and (2) -  that every $a \in M'X$ has a proof w.r.t.  $\cal{P}$ and $M'X$.

Note that it must be the case that $M' \cap X= \emptyset$ and $M' \cap S= \emptyset$.

1.  Assume that $M'X$ is not a 
model of $\cal{P}$. Then, there is a rule $r= H \lpimp \bpos, \bnbd$ in $\cal{P}$ 
such that  $M'X$ satisfies the body of $r$ and the head $H$ has 
empty intersection with $M'X$.
Note that $r$ is not in ${\cal{P}}_S$. Otherwise it would not be violated by 
$M'X$, since $X$ is a model of ${\cal{P}}_S$, no atom in $S$ is in $\cal{P}'$ and $M'$ is a stable model of $\cal{P}'$.

Since the body of $r$  is satisfied by $M'X$ and $M' \cap X= \emptyset$, $\bpos$ can always be written as
$(B_{M'} \cup B_X)$, where $B_{M'} = (B \cap M')$, $B_X = (B \cap X)$, and $B_{M'} \cap B_X =\emptyset$, and $\bnbd$
can always be written as
$(B' \cup B_{S-X})$, where $B_{S-X} = (B \cap (S-X))$, $B'=B- B_{S-X}$,  and $B' \cap (M' \cup S)=\emptyset$.
Analogously,
$H$ can be written as $H' \cup H_{S - X}$, where $H_{S -  X} = (H\cap 
(S - X))$, and $H'=H-H_{S-X}$.

After executing procedure $\mbox{\Reduce}$$({\cal{P}},X,(S-X))$, $\cal{P}'$ will contain the rule 
$r':H' \lpimp \btpos, \btnbd$, where  $\btpos= B_{M'}$ and $\btnbd=B'$. Since the body of $r$ is satisfied by $M'X$, and $M' \cap S=\emptyset$, it must be the case that the body of $r'$ is satisfied by $M'$.  But, since $H$ has an empty intersection with $M'X$ and $H' \subseteq H$, $H'$ has an empty intersection with $M'X$. So it must be the case that $H' \cap M' =\emptyset$.  Thus $r'$ is violated by $M'$, 
and then $M'$ is not a model of $\cal{P}'$ which contradicts the hypothesis. So it must be the case that $M'X$ is a model of $\cal{P}$.

2. Next we show that every $a \in M'X$ has a proof w.r.t.  $\cal{P}$ and $M'X$.  
First, we show that if $a \in X$, then $a$ has a proof w.r.t. $\cal{P}$ and $M'X$. Since $a \in X$ and $X$ is a stable model of ${\cal{P}}_S$, $a$ has a proof w.r.t. ${\cal{P}}_S$ and $X$. The proof is by induction on the length $n$ of the proof of $a$ w.r.t. ${\cal{P}}_S$ and $X$.
\begin{description}
\item[$n=1$] So the proof of $a$ is a single rule $r \in {\cal{P}}_S$ of the form $H \lpimp \bnbd$, where $H\cap X=\{a\}$ and $\bnbd \subseteq S-X$. Since $r\in {\cal{P}}_S$ and no atom in $S$ is in $\cal{P}'$ it must be the case that $\bnbd \cap M'X = \emptyset $ and $H\cap M'X=\{a\}$. In addition, by the definition of  ${\cal{P}}_S$, since $r\in {\cal{P}}_S$ $r\in {\cal{P}}$. So $r$ is the proof of $a$ w.r.t. ${\cal{P}}$ and $M'X$.
\item[$n>1$] We assume that for every $k<n$, if an atom $b$ has a proof  of length $k$  w.r.t. ${\cal{P}}_S$ and $X$, then $b$ has a proof w.r.t. ${\cal{P}}$ and $M'X$. Assume now that for some atom $a \in X$, $a$ has a proof  of length $n$  w.r.t. ${\cal{P}}_S$ and $X$. Let $r \in {\cal{P}}_S$ be the last rule in that proof of $a$. The rule $r$ must be of the form $H \lpimp \bpos, \bnbd$, where $H\cap X=\{a\}$, every $b \in \bpos$ has a proof of legth $k<n$ w.r.t. ${\cal{P}}_S$ and $X$, and $\bnbd \subseteq S-X$. By the induction hypothesis, for every $b \in \bpos$ there is a proof of $b$ w.r.t. ${\cal{P}}$ and $M'X$. Since $r\in {\cal{P}}_S$ and no atom in $S$ is in $\cal{P}'$ it must be the case that $\bnbd \cap M'X = \emptyset $ and $H\cap M'X=\{a\}$. In addition, by the definition of  ${\cal{P}}_S$, since $r\in {\cal{P}}_S$ $r\in {\cal{P}}$. So $r$ is the last rule in a proof of $a$ w.r.t. ${\cal{P}}$ and $M'X$.
\end{description}
Second, we show that if $a \in M'$, then $a$ has a proof w.r.t. $\cal{P}$ and $M'X$. Since $a \in M'$ and $M'$ is a stable model of $\cal{P}'$, $a$ has a proof w.r.t. ${\cal{P}}'$ and $M'$. The proof is by induction on the length $n$ of the proof of $a$ w.r.t. ${\cal{P}}'$ and $M'$.
\begin{description}
\item[$n=1$] So the proof of $a$ is a single rule $r \in {\cal{P}}'$ of the form $H \lpimp \bnbd$, where $H\cap M'=\{a\}$ and $\bnbd \cap M' =\emptyset$. Since $r\in {\cal{P}}'$, and by the way {\it Reduce} works,  there must be a rule $r' \in {\cal{P}}$ such that $r'$ is of the form  $H \cup H_{S-X} \lpimp \bpos, \btnbd$, where $H_{S-X}$ is the set of atoms in the head of $r'$ that belong to $S-X$,
 $\bpos \subseteq X$, and  $\btnbd = \bnbd \cup B_{S-X}$, where $B_{S-X}$ is the set of atoms that appear negative in the body of $r'$ and belong to $S-X$. By Part 1 of this proof, since $\bpos \subseteq X$, every atom in $\bpos$ has a proof  w.r.t. ${\cal{P}}$ and $M'X$. Considering 
$\btnbd = \bnbd \cup B_{S-X}$,  Since $S \cap M' = \emptyset$, $\bnbd \cap M'X=\emptyset$, and $B_{S-X} \cap M'X=\emptyset$, so $\btnbd \cap M'X=\emptyset$. In addition, since $H\cap M'=\{a\}$ and $S \cap M' = \emptyset$, $(H\cup H_{S-X})\cap M'=\{a\}$. 
So $r'$ is the last rule in a proof of $a$ w.r.t. ${\cal{P}}$ and $M'X$, and hence $a$ has a proof w.r.t. ${\cal{P}}$ and $M'X$.

\item[$n>1$] We assume that for every $k<n$, if an atom $b$ has a proof  of length $k$  w.r.t. ${\cal{P}}'$ and $M'$, then $b$ has a proof w.r.t. ${\cal{P}}$ and $M'X$. Assume now that for some atom $a \in M'$, $a$ has a proof  of length $n$  w.r.t. ${\cal{P}}'$ and $M'$. Let $r \in {\cal{P}}'$ be the last rule in that proof of $a$. The rule $r$ must be of the form $H \lpimp \bpos, \bnbd$, where $H\cap M'=\{a\}$, every $b \in \bpos$ has a proof of legth $k<n$ w.r.t. ${\cal{P}}'$ and $M'$, and for each $d \in \bnbd$ $d \notin M'$, and by the way {\it Reduce} works, $d \notin S$ .
Since $r\in {\cal{P}}'$, and by the way {\it Reduce} works,  there must be a rule $r' \in {\cal{P}}$ such that $r'$ is of the form  $H \cup H_{S-X} \lpimp \btpos, \btnbd$, where:
\begin{description}
\item[ $H_{S-X}$ ] is the set of atoms in the head of $r'$ that belong to $S-X$,
\item[$\btpos= \bpos \cup B_X$,] where $B_X$ is a set of atoms that belong to $X$,
\item[$\btnbd= \bnbd \cup B_{S-X}$,] where $B_{S-X}$ is a set of atoms that  belong to $S-X$.
\end{description}
We note that:
\begin{enumerate}
\item By the induction hypothesis, for every $b \in \bpos$ there is a proof of $b$ w.r.t. ${\cal{P}}$ and $M'X$.
\item By Part 1 of this proof, since $B_X \subseteq X$, every atom in $\bpos$ has a proof  w.r.t. ${\cal{P}}$ and $M'X$.
\item Since for each $d \in \bnbd$ $d \notin M'$, and by the way {\it Reduce} works, $d$ is also not in  $ S$ and therefore not in $X$,  $d$ is not in $M'X$.
\item For each $d \in B_{S-X}$ $d$ is not in $X$ and by the way {\it Reduce } works, $d$ is not in $M'$. So for every $d \in B_{S-X}$, $d$ is not in $M'X$.
\item Since $H\cap M'=\{a\}$ and no atoms from $S$ is in $M'$, it must be the case that $(H \cup H_{S-X}) \cap M'X =\{a\}$ .
\end{enumerate}
From all of the above it follows that $r'$ is the last rule of a proof of $a$ w.r.t. ${\cal{P}}$ and $M'X$. 
\end{description}


Consider the program $\cal{P}$ from Example \ref{eg}, which has two stable models: $\{a,c\}$ and $\{b,d\}$. Let us compute the stable models of $\cal{P}$ according to Theorem \ref{gd}. We take
$U=\{a,b\}$, which is a g-splitting set for $\cal{P}$ . The bottom of $\cal{P}$ according to $U$, denoted $b_{\{a,b\}}(\cal{P})$, are Rule 1 and Rule 2, that is: $\{ a \lpimp \nbd b$, 
$b \lpimp\nbd a\}$. So the bottom has two stable models: $\{a\}$, and $\{b\}$. If we propagate the model $\{a\}$ to the top of the program, we are left with the rule $\{c \lpimp \}$, and we get the stable model $\{a,c\}$. If we propagate the model $\{b\}$ to the top of the program, we are left with the rule $\{d \lpimp \}$, and we get the stable model $\{b,d\}$.


\section{Related Work}
\label{related}
The idea of splitting is discussed in many publications. Here we discuss papers that deal with generating splitting sets and relaxing the definition of a splitting set.

The work in \cite{JWHY15} suggests a new way of splitting that introduces a possibly exponential number of new atoms to the program. The authors show that for some typical programs their splitting method is efficient, but clearly it can be quite resource demanding in the worst case.

Baumann \cite{Bau11} discuss  splitting sets and graphs, but they do not go all the way in introducing a polynomial algorithm for computing classical splitting sets, as we do here. The authors of \cite{BBDW12} suggest {\em quasi-splitting}, a relaxation of the concept of splitting that requires the introduction of new atoms to the program, and they describe a polynomial algorithm, based on the dependency graph of the program, to efficiently compute a quasi-splitting set. Our algorithm is essentially a search algorithm with fractions of the dependency graph as states in the search space. We do not need the introduction of new atoms to define g-splitting sets.

\section{Conclusions}\label{sect:conclusions}
The concept of splitting has a considerable role in logic programming. 
This paper has two major contributions. First, we show that the task of looking for an appropriate splitting set can be formulated as a classical search problem and computed in time that is polynomial in the size of the program. Search has been studied extensively in AI, and when we formulate a problem as a search problem, we immediately benefit from the library of search algorithms and strategies that has developed in the past and will be generated in the future. 
Our second contribution is introducing g-splitting sets, which are a generalization of the definition of splitting sets, as presented by Lifschitz and Turner. This allows for a larger set of programs to be split to non-trivial parts.

\bibliographystyle{eptcs}
\bibliography{miscbib,rachel,dis,split}

\begin{thebibliography}{10}
\providecommand{\bibitemdeclare}[2]{}
\providecommand{\surnamestart}{}
\providecommand{\surnameend}{}
\providecommand{\urlprefix}{Available at }
\providecommand{\url}[1]{\texttt{#1}}
\providecommand{\href}[2]{\texttt{#2}}
\providecommand{\urlalt}[2]{\href{#1}{#2}}
\providecommand{\doi}[1]{doi:\urlalt{http://dx.doi.org/#1}{#1}}
\providecommand{\bibinfo}[2]{#2}

\bibitemdeclare{incollection}{Bau11}
\bibitem{Bau11}
\bibinfo{author}{Ringo \surnamestart Baumann\surnameend}
  (\bibinfo{year}{2011}): \emph{\bibinfo{title}{Splitting an Argumentation
  Framework}}.
\newblock In \bibinfo{editor}{James~P. \surnamestart Delgrande\surnameend} \&
  \bibinfo{editor}{Wolfgang \surnamestart Faber\surnameend}, editors: {\sl
  \bibinfo{booktitle}{Logic Programming and Nonmonotonic Reasoning}},
  \bibinfo{publisher}{Springer Berlin Heidelberg}, \bibinfo{address}{Berlin,
  Heidelberg}, pp. \bibinfo{pages}{40--53}, \doi{10.1007/978-3-642-20895-9\_6}.

\bibitemdeclare{incollection}{BBDW12}
\bibitem{BBDW12}
\bibinfo{author}{Ringo \surnamestart Baumann\surnameend},
  \bibinfo{author}{Gerhard \surnamestart Brewka\surnameend},
  \bibinfo{author}{Wolfgang \surnamestart Dvo{\v{r}}{\'a}k\surnameend} \&
  \bibinfo{author}{Stefan \surnamestart Woltran\surnameend}
  (\bibinfo{year}{2012}): \emph{\bibinfo{title}{Parameterized Splitting: A
  Simple Modification-Based Approach}}.
\newblock In \bibinfo{editor}{Esra \surnamestart Erdem\surnameend},
  \bibinfo{editor}{Joohyung \surnamestart Lee\surnameend},
  \bibinfo{editor}{Yuliya \surnamestart Lierler\surnameend} \&
  \bibinfo{editor}{David \surnamestart Pearce\surnameend}, editors: {\sl
  \bibinfo{booktitle}{Correct Reasoning: Essays on Logic-Based AI in Honour of
  Vladimir Lifschitz}}, \bibinfo{publisher}{Springer Berlin Heidelberg},
  \bibinfo{address}{Berlin, Heidelberg}, pp. \bibinfo{pages}{57--71},
  \doi{10.1007/978-3-642-30743-0_5}.

\bibitemdeclare{article}{BeDe94}
\bibitem{BeDe94}
\bibinfo{author}{Rachel \surnamestart Ben-Eliyahu\surnameend} \&
  \bibinfo{author}{Rina \surnamestart Dechter\surnameend}
  (\bibinfo{year}{1994}): \emph{\bibinfo{title}{Propositional Semantics For
  Disjunctive Logic Programs}}.
\newblock {\sl \bibinfo{journal}{Annals of Mathematics and Artificial
  Intelligence}} \bibinfo{volume}{12}, pp. \bibinfo{pages}{53--87},
  \doi{10.1007/BF01530761}.

\bibitemdeclare{inproceedings}{DEFK09}
\bibitem{DEFK09}
\bibinfo{author}{Minh \surnamestart Dao-Tran\surnameend},
  \bibinfo{author}{Thomas \surnamestart Eiter\surnameend},
  \bibinfo{author}{Michael \surnamestart Fink\surnameend} \&
  \bibinfo{author}{Thomas \surnamestart Krennwallner\surnameend}
  (\bibinfo{year}{2009}): \emph{\bibinfo{title}{Modular Nonmonotonic Logic
  Programming Revisited}}.
\newblock In \bibinfo{editor}{Patricia~M. \surnamestart Hill\surnameend} \&
  \bibinfo{editor}{David~S. \surnamestart Warren\surnameend}, editors: {\sl
  \bibinfo{booktitle}{Logic Programming}}, \bibinfo{publisher}{Springer Berlin
  Heidelberg}, \bibinfo{address}{Berlin, Heidelberg}, pp.
  \bibinfo{pages}{145--159}, \doi{10.1007/978-3-642-02846-5\_16}.

\bibitemdeclare{article}{DLL62}
\bibitem{DLL62}
\bibinfo{author}{Martin \surnamestart Davis\surnameend},
  \bibinfo{author}{George \surnamestart Logemann\surnameend} \&
  \bibinfo{author}{Donald \surnamestart Loveland\surnameend}
  (\bibinfo{year}{1962}): \emph{\bibinfo{title}{A machine program for
  theorem-proving}}.
\newblock {\sl \bibinfo{journal}{Communications of the ACM}}
  \bibinfo{volume}{5}(\bibinfo{number}{7}), pp. \bibinfo{pages}{394--397},
  \doi{10.1145/368273.368557}.

\bibitemdeclare{book}{Dec03}
\bibitem{Dec03}
\bibinfo{author}{Rina \surnamestart Dechter\surnameend} (\bibinfo{year}{2003}):
  \emph{\bibinfo{title}{Constraint processing}}.
\newblock \bibinfo{publisher}{Morgan Kaufmann}.

\bibitemdeclare{article}{Dij59}
\bibitem{Dij59}
\bibinfo{author}{Edsger~W. \surnamestart Dijkstra\surnameend}
  (\bibinfo{year}{1959}): \emph{\bibinfo{title}{A note on two problems in
  connexion with graphs}}.
\newblock {\sl \bibinfo{journal}{Numerische mathematik}}
  \bibinfo{volume}{1}(\bibinfo{number}{1}), pp. \bibinfo{pages}{269--271},
  \doi{10.1007/BF01386390}.

\bibitemdeclare{inproceedings}{Fel11}
\bibitem{Fel11}
\bibinfo{author}{Ariel \surnamestart Felner\surnameend} (\bibinfo{year}{2011}):
  \emph{\bibinfo{title}{Position paper: Dijkstra's algorithm versus uniform
  cost search or a case against dijkstra's algorithm}}.
\newblock In: {\sl \bibinfo{booktitle}{Fourth annual symposium on combinatorial
  search}}, pp. \bibinfo{pages}{47--51}.

\bibitemdeclare{inproceedings}{FLLP09}
\bibitem{FLLP09}
\bibinfo{author}{Paolo \surnamestart Ferraris\surnameend},
  \bibinfo{author}{Joohyung \surnamestart Lee\surnameend},
  \bibinfo{author}{Vladimir \surnamestart Lifschitz\surnameend} \&
  \bibinfo{author}{Ravi \surnamestart Palla\surnameend} (\bibinfo{year}{2009}):
  \emph{\bibinfo{title}{Symmetric splitting in the general theory of stable
  models}}.
\newblock In: {\sl \bibinfo{booktitle}{Twenty-First International Joint
  Conference on Artificial Intelligence}}, pp. \bibinfo{pages}{797--803}.

\bibitemdeclare{inproceedings}{GKKOST08}
\bibitem{GKKOST08}
\bibinfo{author}{Martin \surnamestart Gebser\surnameend},
  \bibinfo{author}{Roland \surnamestart Kaminski\surnameend},
  \bibinfo{author}{Benjamin \surnamestart Kaufmann\surnameend},
  \bibinfo{author}{Max \surnamestart Ostrowski\surnameend},
  \bibinfo{author}{Torsten \surnamestart Schaub\surnameend} \&
  \bibinfo{author}{Sven \surnamestart Thiele\surnameend}
  (\bibinfo{year}{2008}): \emph{\bibinfo{title}{Engineering an Incremental ASP
  Solver}}.
\newblock In \bibinfo{editor}{Maria \surnamestart Garcia de~la
  Banda\surnameend} \& \bibinfo{editor}{Enrico \surnamestart
  Pontelli\surnameend}, editors: {\sl \bibinfo{booktitle}{Logic Programming}},
  \bibinfo{publisher}{Springer Berlin Heidelberg}, \bibinfo{address}{Berlin,
  Heidelberg}, pp. \bibinfo{pages}{190--205},
  \doi{10.1007/978-3-540-89982-2\_23}.

\bibitemdeclare{article}{GeLi91}
\bibitem{GeLi91}
\bibinfo{author}{Michael \surnamestart Gelfond\surnameend} \&
  \bibinfo{author}{Vladimir \surnamestart Lifschitz\surnameend}
  (\bibinfo{year}{1991}): \emph{\bibinfo{title}{Classical Negation in Logic
  Programs and Disjunctive Databases}}.
\newblock {\sl \bibinfo{journal}{New Generation Computing}}
  \bibinfo{volume}{9}, pp. \bibinfo{pages}{365--385}, \doi{10.1007/BF03037169}.

\bibitemdeclare{article}{JOTW09}
\bibitem{JOTW09}
\bibinfo{author}{Tomi \surnamestart Janhunen\surnameend},
  \bibinfo{author}{Emilia \surnamestart Oikarinen\surnameend},
  \bibinfo{author}{Hans \surnamestart Tompits\surnameend} \&
  \bibinfo{author}{Stefan \surnamestart Woltran\surnameend}
  (\bibinfo{year}{2009}): \emph{\bibinfo{title}{Modularity aspects of
  disjunctive stable models}}.
\newblock {\sl \bibinfo{journal}{Journal of Artificial Intelligence Research}}
  \bibinfo{volume}{35}, pp. \bibinfo{pages}{813--857}, \doi{10.1613/jair.2810}.

\bibitemdeclare{inproceedings}{JWHY15}
\bibitem{JWHY15}
\bibinfo{author}{Jianmin \surnamestart Ji\surnameend}, \bibinfo{author}{Hai
  \surnamestart Wan\surnameend}, \bibinfo{author}{Ziwei \surnamestart
  Huo\surnameend} \& \bibinfo{author}{Zhenfeng \surnamestart Yuan\surnameend}
  (\bibinfo{year}{2015}): \emph{\bibinfo{title}{Splitting a Logic Program
  Revisited}}.
\newblock In: {\sl \bibinfo{booktitle}{Proceedings of the Twenty-Ninth AAAI
  Conference on Artificial Intelligence}}, \bibinfo{series}{AAAI'15},
  \bibinfo{publisher}{AAAI Press}, pp. \bibinfo{pages}{1511--1517}.
\newblock \urlprefix\url{http://dl.acm.org/citation.cfm?id=2886521.2886530}.

\bibitemdeclare{inproceedings}{LiTu94}
\bibitem{LiTu94}
\bibinfo{author}{Vladimir \surnamestart Lifschitz\surnameend} \&
  \bibinfo{author}{Hudson \surnamestart Turner\surnameend}
  (\bibinfo{year}{1994}): \emph{\bibinfo{title}{Splitting a Logic Program.}}
\newblock In: {\sl \bibinfo{booktitle}{ICLP}}, \bibinfo{volume}{94}, pp.
  \bibinfo{pages}{23--37}.

\bibitemdeclare{article}{OiJa08}
\bibitem{OiJa08}
\bibinfo{author}{Emilia \surnamestart Oikarinen\surnameend} \&
  \bibinfo{author}{Tomi \surnamestart Janhunen\surnameend}
  (\bibinfo{year}{2008}): \emph{\bibinfo{title}{Achieving compositionality of
  the stable model semantics for smodels programs}}.
\newblock {\sl \bibinfo{journal}{Theory and Practice of Logic Programming}}
  \bibinfo{volume}{8}(\bibinfo{number}{5-6}), p. \bibinfo{pages}{717–761},
  \doi{10.1017/S147106840800358X}.

\bibitemdeclare{book}{Pea84}
\bibitem{Pea84}
\bibinfo{author}{Judea \surnamestart Pearl\surnameend} (\bibinfo{year}{1984}):
  \emph{\bibinfo{title}{Heuristics: intelligent search strategies for computer
  problem solving}}.
\newblock \bibinfo{publisher}{Addison-Wesley Pub. Co., Inc., Reading, MA}.

\bibitemdeclare{book}{RuNo10}
\bibitem{RuNo10}
\bibinfo{author}{Stuart~J. \surnamestart Russell\surnameend} \&
  \bibinfo{author}{Peter \surnamestart Norvig\surnameend}
  (\bibinfo{year}{2010}): \emph{\bibinfo{title}{Artificial Intelligence - {A}
  Modern Approach, Third International Edition}}.
\newblock \bibinfo{publisher}{Pearson Education}.
\newblock
  \urlprefix\url{http://vig.pearsoned.com/store/product/1,1207,store-12521\_isbn-0136042597,00.html}.

\bibitemdeclare{article}{SeMiLe96}
\bibitem{SeMiLe96}
\bibinfo{author}{Bart \surnamestart Selman\surnameend},
  \bibinfo{author}{David~G \surnamestart Mitchell\surnameend} \&
  \bibinfo{author}{Hector~J \surnamestart Levesque\surnameend}
  (\bibinfo{year}{1996}): \emph{\bibinfo{title}{Generating hard satisfiability
  problems}}.
\newblock {\sl \bibinfo{journal}{Artificial intelligence}}
  \bibinfo{volume}{81}(\bibinfo{number}{1-2}), pp. \bibinfo{pages}{17--29},
  \doi{10.1016/0004-3702(95)00045-3}.

\end{thebibliography}
\end{document}